\documentclass[10pt,twocolumn,letterpaper]{article}

\usepackage{cvpr}
\usepackage{times}
\usepackage{epsfig}
\usepackage[font=small,skip=5pt]{caption}
\usepackage{graphicx}
\usepackage{amsmath}
\usepackage{amssymb}
\usepackage{comment}

\newcommand{\bp}{\textbf{p}}
\newcommand{\bb}{\textbf{b}}
\newcommand{\bx}{\textbf{x}}

\newcommand{\bA}{\textbf{A}}

\newcommand{\bF}{\textbf{F}}
\newcommand{\bG}{\textbf{G}}
\newcommand{\bT}{\textbf{T}}

\newcommand{\aaron}[1]{{{\bf[A: #1]}}}

\graphicspath{{images/}}

\usepackage[breaklinks=true,bookmarks=false]{hyperref}

\cvprfinalcopy 

\def\cvprPaperID{1592} 

\ifcvprfinal\fi
\begin{document}


\title{Controlling Perceptual Factors in Neural Style Transfer}
\author{Leon A. Gatys${}^1$\quad
Alexander S. Ecker${}^1$\quad
Matthias Bethge${}^1$\quad
Aaron Hertzmann${}^2$\quad
Eli Shechtman${}^2$ \\
${}^1$University of T\"ubingen\qquad
${}^2$Adobe Research}

\twocolumn[{%
\renewcommand\twocolumn[1][]{#1}%
\maketitle
\begin{center}
\includegraphics[width=1\textwidth]{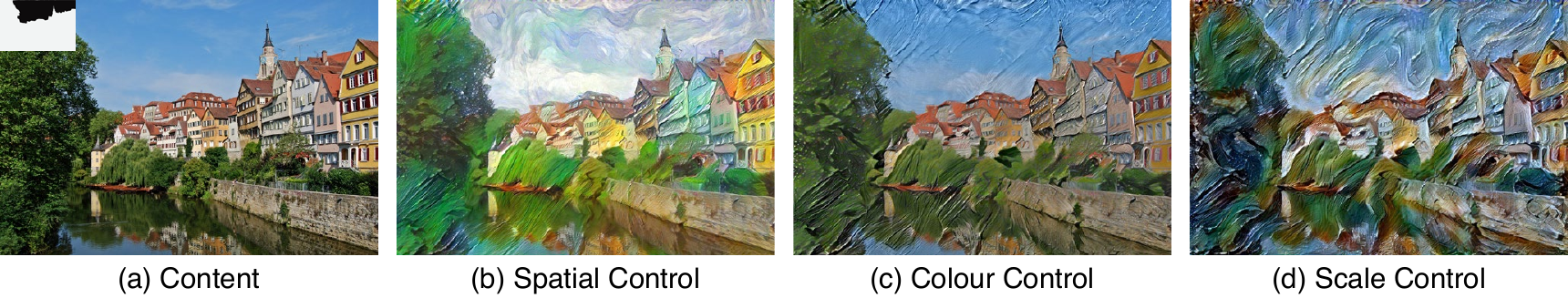}
\captionof{figure}{
Overview of our control methods.
\textbf{(a)} Content image, with spatial mask inset. 
\textbf{(b)} Spatial Control. The sky is stylised using the sky of Style II from Fig.~\ref{fig:SpatialControl}(c). The ground is stylised using Style I from Fig.~\ref{fig:ScaleControl}(b).  
\textbf{(c)} Colour Control. The colour of the content image is preserved using luminance-only style transfer described in Section \ref{sec:LuminanceTransfer}. 
\textbf{(d)} Scale Control. The fine scale is stylised using using Style I from Fig.~\ref{fig:ScaleControl}(b) and the coarse scale is stylised using Style III from Fig.~\ref{fig:ScaleControl}(b). Colour is preserved using the colour matching described in section \ref{sec:ColourMatching}.
}
\label{fig:Teaser}
\end{center}
}]

\begin{abstract}
Neural Style Transfer has shown very exciting results enabling new forms of image manipulation. Here we extend the existing method to introduce control over spatial location, colour information and across spatial scale\footnote{Code: github.com/leongatys/NeuralImageSynthesis}\footnote{Supplement: bethgelab.org/media/uploads/stylecontrol/supplement/}. We demonstrate how this enhances the method by allowing high-resolution controlled stylisation and helps to alleviate common failure cases such as applying ground textures to sky regions. Furthermore, by decomposing style into these perceptual factors we enable the combination of style information from multiple sources to generate new, perceptually appealing styles from existing ones. We also describe how these methods can be used to more efficiently produce large size, high-quality stylisation. Finally we show how the introduced control measures can be applied in recent methods for Fast Neural Style Transfer. 
\end{abstract}

\section{Introduction}
Example-based style transfer is a major way to create new, perceptually appealing images from existing ones.
It takes two images $\bx_S$ and $\bx_C$ as input, and produces a new image $\hat{\bx}$ applying the style of $\bx_S$ to the content of $\bx_C$. The concepts of ``style" and ``content" are both expressed in terms of image statistics; for example, two images are said to have the same style if they embody the same correlations of specific image features. To provide intuitive control over this process, one must identify ways to access perceptual factors in these statistics.

In order to identify these factors, we observe some of the different ways that one might describe an artwork such as Vincent van Gogh's \textit{A Wheatfield with Cypresses} (Fig.~\ref{fig:SpatialControl}(c)). First, one might separately describe different styles in different regions, such as in the sky as compared to the ground. Second, one might describe the colour palette, and how it relates to the underlying scene, separately from factors like image composition or brush stroke texture. Third, one might describe fine-scale spatial structures, such as brush stroke shape and texture, separately from coarse-scale structures like the arrangements of strokes and the swirly structure in the sky of the painting. These observation motivates our hypothesis: image style can be perceptually factorised into style in different spatial regions, colour and luminance information, and across spatial scales, making them meaningful control dimensions for image stylisation.

Here we build on this hypothesis to introduce meaningful control to a recent image stylisation method known as Neural Style Transfer \cite{gatys} in which the image statistics that capture content and style are defined on feature responses in a Convolutional Neural Network (CNN) \cite{VGG}. Namely, we introduce methods for controlling image stylisation independently in different spatial regions (Fig.~\ref{fig:Teaser}(b)), for colour and luminance information (Fig.~\ref{fig:Teaser}(c)) as well as on different spatial scales (Fig.~\ref{fig:Teaser}(d)). We show how they can be applied to improve Neural Style Transfer and to alleviate some of its common failure cases. Moreover, we demonstrate how the factorisation of style into these aspects can gracefully combine style information from multiple images and thus enable the creation of new, perceptually interesting styles. 
We also show a method for efficiently rendering high-resolution stylisations using a coarse-to-fine approach that reduced optimisation time by an approximate factor of $2.5$.
Finally, we show that in addition to the original optimisation-based style transfer, these control methods can also be applied to recent fast approximations of Neural Style Transfer \cite{johnson_perceptual_2016, ulyanov_texture_2016}

\section{Related Work} 

There is a large body of work on image stylisation techniques. The first example-based technique was Image Analogies \cite{image-analogies}, which built on patch-based texture synthesis techniques \cite{efros-leung, wei-levoy}. This method introduced stylisation based on an example painting, as well as ways to preserve colour, and to control stylisation of different regions separately. The method used a coarse-to-fine texture synthesis procedure for speed \cite{wei-levoy}. Since then, improvements to the optimisation method and new applications \cite{ramanarayan-bala,stylit} have been proposed. 
Patch-based methods have also been used with CNN features \cite{liWand2016MRF, champandard_semantic_2016}, leading to improved texture representations and stylisation results. 
Scale control has been developed for patch-based texture synthesis \cite{HRRG08} and many other techniques have been developed for transferring colour style \cite{CGF:CGF12671}.
There are also many procedural stylisation techniques that provide extensive user control in the non-photorealistic rendering literature, e.g., \cite{painting-with-bob,wysiwygnpr,anipaint}. These procedural methods provide separate controls for adjusting spatial variation in styles, colour transformation, and brush stroke style, but cannot work from training data.

More recently, Neural Style Transfer \cite{gatys} has demonstrated impressive results in example-based image stylisation. The method is based on a parametric texture model \cite{julesz_visual_1962, heeger_pyramid-based_1995, portilla_parametric_2000} defined by summary statistics on CNN responses \cite{gatys_texture_2015} and appears to have several advantages over patch-based synthesis. Most prominently, during the stylisation it displays a greater flexibility to create new image structures that are not already present in the source images \cite{liWand2016MRF}.

However, the representation of image style within the parametric neural texture model \cite{gatys_texture_2015} allows far less intuitive control over the stylisation outcome than patch-based methods. The texture parameters can be used to influence the stylisation but their interplay is extremely complex due to the complexity of the deep representations they are defined on. Therefore it is difficult to predict their perceptual effect on the stylisation result.
Our main goal in this work is to introduce intuitive ways to control Neural Style Transfer to combine the advantages of that method with the more fine-grained user control of earlier stylisation methods. Note that concurrent work \cite{wilmot_stable_2017} independently developed a similar approach for spatial control as presented here.
 
\section{Neural Style Transfer}

The Neural Style Transfer method \cite{gatys} works as follows.
We define a content image $\bx_C$ and a style image $\bx_S$ with corresponding feature representations $\bF_\ell(\bx_C)$ and $\bF_\ell(\bx_S)$ in layer $\ell$ of a CNN. Each column of $\bF_\ell(\bx)$ is a vectorised feature map and thus  $\bF_\ell \in \mathcal{R}^{M_\ell(\bx)\times N_\ell}$ where $N_\ell$ is the number of feature maps in layer $\ell$ and $M_\ell(\bx) = H_\ell(\bx)\times W_\ell(\bx)$ is the product of height and width of each feature map. Note that while $N_\ell$ is independent of the input image, $M_\ell(\bx)$ depends on the size of the input image. 

Neural Style Transfer generates a new image $\hat{\bx}$ that depicts the content of image $\bx_C$ in the style of image $\bx_S$ by minimising following loss function with respect to $\hat{\bx}$ 
\begin{align}
\mathcal{L}_{total} = \alpha \mathcal{L}_{content} + \beta \mathcal{L}_{style} 
\end{align}
where the content term compares feature maps at a single layer $\ell_C$:
\begin{align}
\mathcal{L}_{content} = \frac{1}{N_{\ell_{c}}M_{\ell_{c}}(\bx_C)} \sum_{ij}\left(\bF_{\ell_{c}}(\hat{\bx})-\bF_{\ell_{c}}(\bx_C)\right)_{ij}^2
\end{align}
and the style term compares a set of summary statistics:
\begin{align}
\label{eq:opt}
\mathcal{L}_{style} &= \sum_\ell w_\ell E_\ell \\
E_\ell &= \frac{1}{4N_\ell^2}\sum_{ij}{\left(\bG_\ell(\hat{\bx}) - \bG_\ell(\bx_S)\right)_{ij}^{2}}
\end{align}
where $\bG_\ell(\bx)=\frac{1}{M_\ell(\bx)}\bF_\ell(\bx)^T\bF_\ell(\bx)$ is the Gram Matrix of the feature maps in layer $\ell$ in response to image $\bx$.
As in the original work \cite{gatys}, we use the VGG-19 Network and include ``conv4\_2'' as the layer $\ell_C$ for the image content and Gram Matrices from layers  ``conv1\_1'',``conv2\_1'',``conv3\_1'',``conv4\_1'',``conv5\_1'' as the image statistics that model style.

\section{Spatial Control}

We first introduce ways to spatially control Neural Style Transfer. Our goal is to control which region of the style image is used to stylise each region in the content image. For example, we would like to apply one style to the sky region and another to the ground region of an image to either avoid artefacts (Fig.~\ref{fig:SpatialControl}(d),(e)) or to generate new combinations of styles from multiple sources (Fig.~\ref{fig:SpatialControl}(f)). 
We take as input $R$ spatial guidance channels $\bT^r$ for both the content and style image (small insets in (Fig.~\ref{fig:SpatialControl}(a)-(c)). 
Each of these is an image map of values in $[0,1]$ specifying which styles should be applied where: regions where the $r^{th}$ content guidance channel is equal to $1$ should get the style from regions where the $r^{th}$ style guidance channel is $1$.
When there are multiple style images, the regions index over all the example images. The guidance channels are propagated to the CNN to produce guidance channels $\bT^r_\ell$ for each layer. This can be done by simple re-sampling or more involved methods as we explain later in this section. We first discuss algorithms for synthesis given the guidance maps.

\subsection{Guided Gram Matrices}\label{sec:GuidedGram}

In the first method we propose, we multiply the feature maps of each layer included in the style features with $R$ guidance channels $\bT^r_\ell$ and compute one spatially guided Gram Matrix for each of the $R$ regions in the style image. 
Formally we define a spatially guided feature map as 
\begin{align}
\bF^r_\ell(\bx)_{[:,i]} = \bT^r_\ell\circ \bF_\ell(\bx)_{[:,i]}
\end{align}
Here $\bF^r_\ell(\bx)_{[:,i]}$ is the i\textsuperscript{th} column vector of $\bF^r_\ell(\bx)$, $r\in R$ and $\circ$ denotes element-wise multiplication. The guidance channel $\bT^r_\ell$ is vectorised and can be either a binary mask for hard guidance or real-valued for soft guidance. 
We normalise $\bT^r_\ell$ such that $\sum_{i}(\bT^r_{\ell})_{i}^2 = 1$. The guided Gram Matrix is then 
\begin{align}
\bG^r_\ell(\bx) = \bF^r_\ell(\bx)^{T}\bF^r_\ell(\bx)
\end{align}
Each guided Gram Matrix is used as the optimisation target for the corresponding region of the content image.
The contribution of layer $\ell$ to the style loss is then:
\begin{align}
\label{eq:guided_loss}
E_\ell = \frac{1}{4N_\ell^2}\sum_{r=1}^{R}{\sum_{ij}{\lambda_{r}\left(\bG^{r}_{\ell}(\hat{\bx}) - \bG^{r}_{\ell}(\bx_S)\right)_{ij}^{2}}}
\end{align}
where $\lambda_r$ is a weighting factor that controls the stylisation strength in the corresponding region $r$. 

\begin{figure}
\includegraphics[width=1\linewidth]{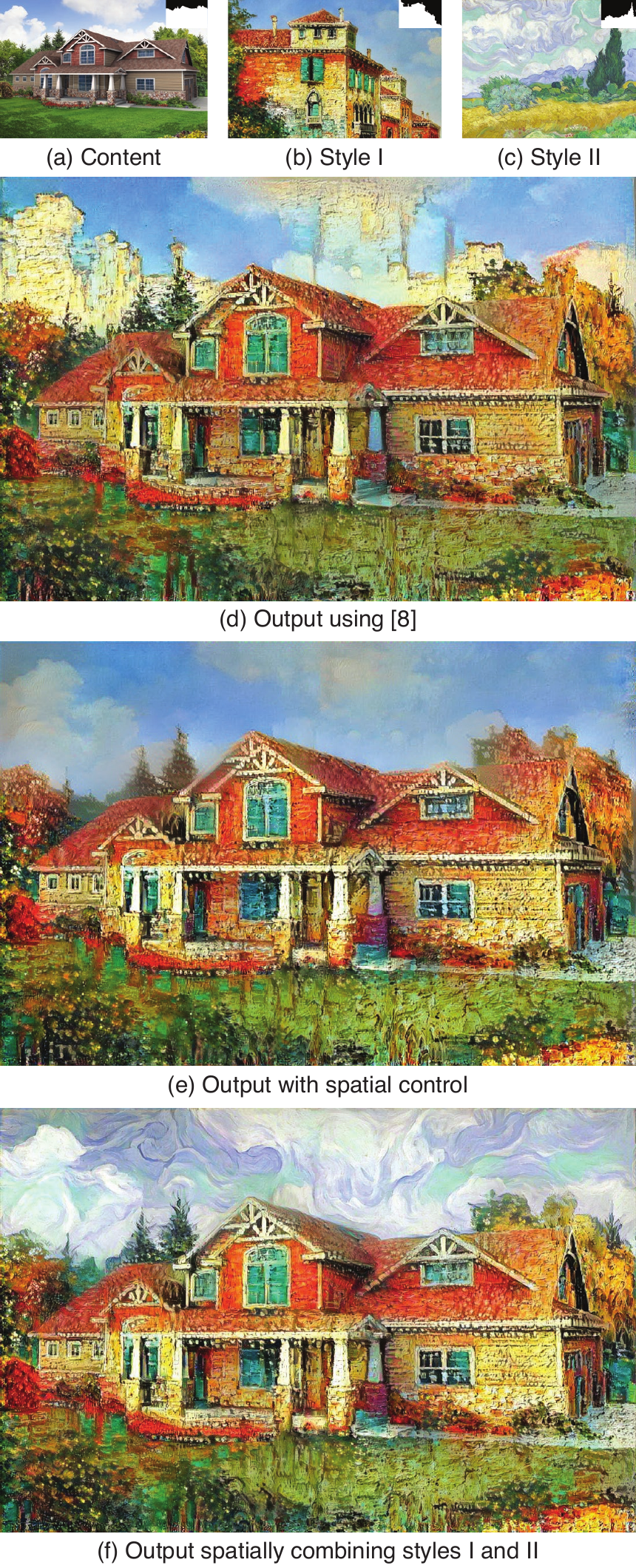}
\caption{
Spatial guidance in Neural Style Transfer.
\textbf{(a)} Content image. 
\textbf{(b)} Style image I. 
\textbf{(c)} Style image II. 
Spatial mask separating the image in sky and ground is shown in the top right corner.
\textbf{(d)} Output from Neural Style Transfer without spatial control \cite{gatys}. The clouds are stylised with image structures from the ground.
\textbf{(e)} Output with spatial guidance.
\textbf{(f)} Output from spatially combining the the ground-style from \textbf{(b)} and the sky-style from \textbf{(c)}.
}
\label{fig:SpatialControl}
\end{figure}

An important use for guidance channels is to ensure that style is transferred between regions of similar scene content in the content and style image.
For example, Figure \ref{fig:SpatialControl} shows an example in which the sky in the content image has bright clouds, whereas the sky in the style image has greyish clouds; as a result, the original style transfer stylises the sky with a bright part of the ground that does not match the appearance of the sky. We address this by dividing both images into a sky and a ground region (Fig.~\ref{fig:SpatialControl}(a),(b) small insets) and require that the sky and ground regions from the painting are used to stylise the respective regions in the photograph (Fig.~\ref{fig:SpatialControl}(e)). 

Given the input guidance channel $\bT^r$, we need to first propagate this channel to produce guidance channels $\bT^r_\ell$ for each layer. The most obvious approach would be to down-sample $\bT^r$ to the dimensions of each layer's feature map.  However, we often find that doing so fails to keep the desired separation of styles by region, e.g., ground texture still appears in the sky. This is because neurons near the boundaries of a guidance region can have large receptive fields that overlap into the other region. 
Instead we use an eroded version of the spatial guiding channels. We enforce spatial guidance only on the neurons whose receptive field is entirely inside the guidance region and add another global guidance channel that is constant over the entire image. We found that this soft spatial guidance usually yields better results. 
For further details on the creation of guidance channels, see the Supplementary Material, section 1.1.

Another application of this method is to generate a new style by combining the styles from multiple example images.  Figure \ref{fig:SpatialControl}(f) shows an example in which the region guidance is used to use the sky style from one image and the ground style from another. This example demonstrates the potential of spatial guidance to combine many example styles together to produce new stylisations.

\subsection{Guided Sums}
Alternatively, instead of computing a Gram Matrix for each guidance channel, we can also just stack the guidance channels with the feature maps as it is done in \cite{champandard_semantic_2016} to spatially guide \textit{neural patches} \cite{liWand2016MRF}. The feature representation of image $\bx$ in layer $\ell$ is then $\bF^\prime_\ell(\bx) = \left[\bF_\ell(\bx),\bT^1_\ell,\bT^2_\ell,...,\bT^R_\ell\right]$ and $\bF^\prime_\ell(\bx) \in \mathcal{R}^{(N_\ell+R)\times M_\ell(\bx)}$. Now the Gram Matrix $\bG^\prime_\ell(\bx)= \frac{1}{M_\ell(\bx)}\bF^\prime_\ell(\bx)^T \bF^\prime_\ell(\bx)$ includes correlations of the image features with the non-zero entries of the guidance channels and therefore encourages that the features in region $r$ of the style image are used to stylise region $r$ in the content image. The contribution of layer $\ell$ to the style loss is simply
\begin{align}
E_\ell = \frac{1}{4N_\ell^2}\sum_{ij}{\left(\bG^\prime_\ell(\hat{\bx}) - \bG^\prime_\ell(\bx_S)\right)_{ij}^{2}}
\end{align}
This is clearly more efficient than the method presented in Section \ref{sec:GuidedGram}. Instead of computing and matching $R$ Gram Matrices one only has to compute one Gram Matrix with $R$ additional channels. Nevertheless, this gain in efficiency comes at the expense of texture quality. The additional channels in the new Gram Matrix are the sums over each feature map spatially weighted by the guidance channel.
\begin{align}
\bG^\prime_\ell(\bx_S)_{i, N_{\ell}+r}= \sum_{j} \left(\bT^r_\ell \circ \bF_\ell(\bx_S)_{[:, i]}\right)_j
\end{align}
Hence this method actually interpolates between matching the original global Gram Matrix stylisation and the spatially weighted sums over the feature maps. While the feature map sums also give a non-trivial texture model, their capacity to model complex textures is limited \cite{gatys_texture_2015}. In practice we find that this method can often give decent results but also does not quite capture the texture of the style image -- as would be expected from the inferior texture model. Results and comparisons can be found in the Supplementary Material, section 1.2.

\section{Colour Control}

\begin{figure}
\includegraphics[width=1\linewidth]{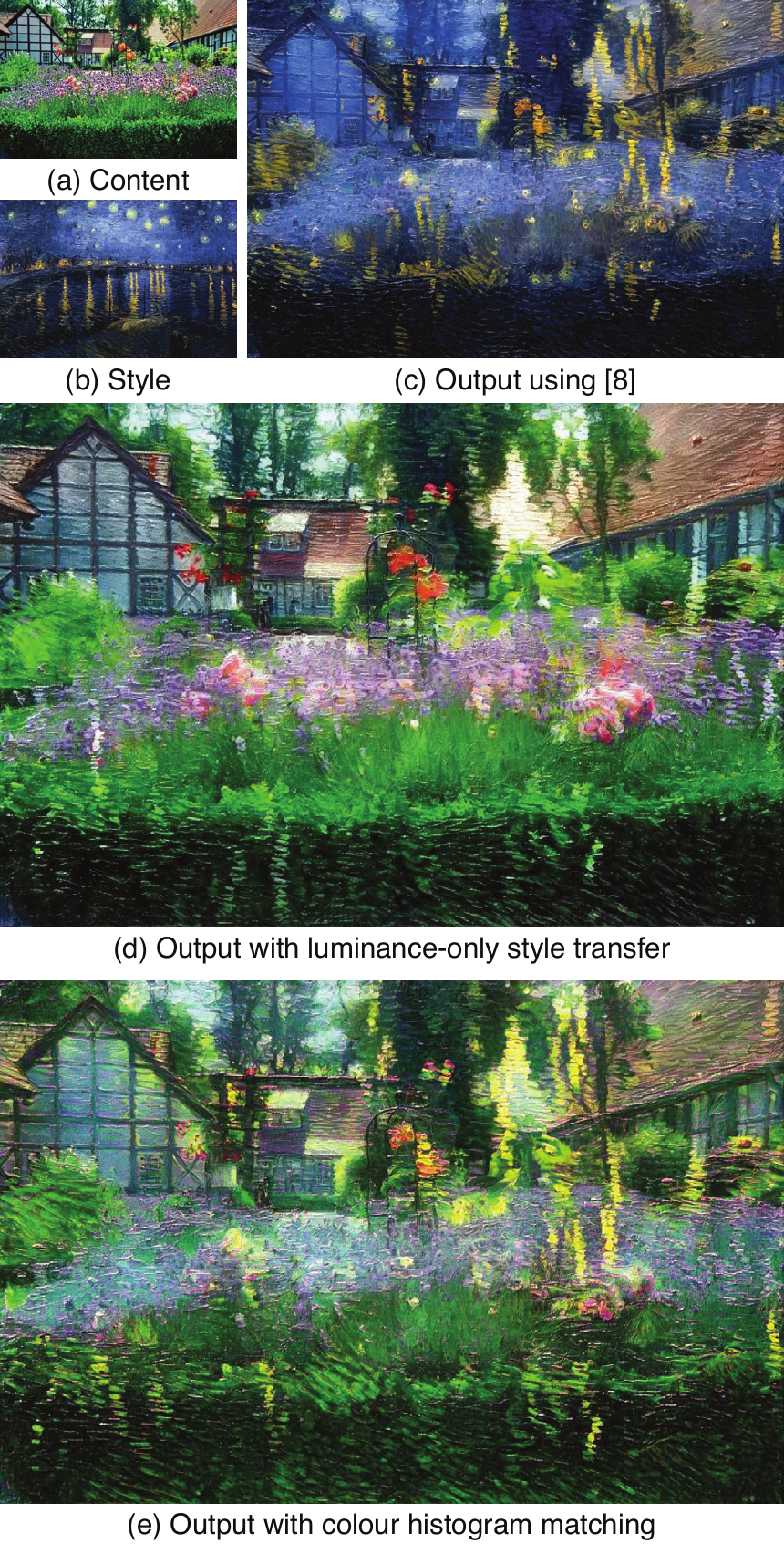}
\caption{Colour preservation in Neural Style Transfer.
\textbf{(a)} Content image.
\textbf{(b)} Style image.
\textbf{(c)} Output from Neural Style Transfer \cite{gatys}. The colour scheme is copied from the painting.
\textbf{(d)} Output using style transfer in luminance domain to preserve colours.
\textbf{(e)} Output using colour transfer to preserve colours.
}
\label{fig:ColourControl}
\end{figure}

The colour information of an image is an important perceptual aspect of its style. At the same time it is largely independent of other style aspects such as the type of brush strokes used or dominating geometric shapes. Therefore it is desirable to independently control the colour information in Neural Style Transfer. A prominent use case for such control is colour preservation during style transfer. When stylising an image using Neural Style Transfer, the output also copies the colour distribution of the style image, which might be undesirable in many cases (Fig.~\ref{fig:ColourControl}(c)).
For example, the stylised farmhouse has the colours of the original van Gogh painting (Fig.~\ref{fig:ColourControl}(c)), whereas one might prefer the output painting to preserve the colours of the farmhouse photograph. In particular, one might imagine that the artist would have used the colours of the scene if they were to paint the farmhouse. Here we present two simple methods to preserve the colours of the source image during Neural Style Transfer --- in other words, to transfer the style without transferring the colours. We compare two different approaches to colour preservation: colour histogram matching and luminance-only transfer (Fig.~\ref{fig:ColourControl}(d,e)).

\subsection{Luminance-only transfer}\label{sec:LuminanceTransfer}
In the first method we perform style transfer only in the luminance channel, as done in Image Analogies \cite{image-analogies}. This is motivated by the observation that visual perception is far more sensitive to changes in luminance than in colour \cite{wandell}.

The modification is simple. The luminance channels $L_S$ and $L_C$ are first extracted from the style and content images. 
Then the Neural Style Transfer algorithm is applied to these images to produce an output luminance image $\hat{L}$.
Using a colour space that separates luminance and colour information, the colour information of the content image is combined with $\hat{L}$ to produce the final colour output image (Fig.~\ref{fig:ColourControl}(d)).

If there is a substantial mismatch between the luminance histogram of the style and the content image, it can be helpful to match the histogram of the style luminance channel $L_S$ to that of the content image $L_C$ before transferring the style. 
For that we simply match mean and variance of the content luminance.  Let $\mu_S$  and $\mu_C$ be the mean luminances of the two images, and $\sigma_S$ and $\sigma_C$ be their standard deviations. Then each luminance pixel in the style image is updated as:
\begin{equation}
L_{s'} = \frac{\sigma_C}{\sigma_S} (L_S - \mu_S) + \mu_C
\label{eq:luminanceNormalization}
\end{equation}

\subsection{Colour histogram matching}\label{sec:ColourMatching}
The second method we present works as follows.
Given the style image $\bx_S$, and the content image $\bx_C$,
the style image's colours are transformed to match the colours of the content image. 
This produces a new style image $\bx_S'$ that replaces $\bx_S$ as input to the Neural Style Transfer algorithm. The algorithm is otherwise unchanged.

The one choice to be made is the colour transfer procedure.
There are many colour transformation algorithms to choose from; see \cite{CGF:CGF12671} for a survey. 
Here we use linear methods, which are simple and effective for colour style transfer. 

Given the style image, each RGB pixel $\bp_S$ is transformed as:
\begin{equation}
\bp_{S}' = \bA \bp_S + \bb
\end{equation}
where $\bA$ is a $3\times3$ matrix and $\bb$ is a 3-vector. 
This transformation is chosen so that the mean and covariance of the RGB values in the new style image $\bp_S'$ match those of $\bp_C'$ \cite{hertzmann-phd} (Appendix B). 
In general, we find that the colour matching method works reasonably well with Neural Style Transfer (Fig.~\ref{fig:ColourControl}(e)), whereas gave poor synthesis results for Image Analogies \cite{hertzmann-phd}.
Furthermore, the colour histogram matching method can also be used to better preserve the colours of the style image. This can substantially improve results for cases in which there is a strong mismatch in colour but one rather wants to keep the colour distribution of the style image (for example with pencil drawings or line art styles). Examples of this application can be found in the Supplementary Material, section 2.2.
 
 \subsection{Comparison}
In conclusion, both methods give perceptually-interesting results but have different advantages and disadvantages. 
The colour-matching method is naturally limited by how well the colour transfer from the content image onto the style image works. The colour distribution often cannot be matched perfectly, leading to a mismatch between the colours of the output image and that of the content image. 

In contrast, the luminance-only transfer method  preserves the colours of the content image perfectly. However, dependencies between the luminance and the colour channels are lost in the output image. While we found that this is usually very difficult to spot, it can be a problem for styles with prominent brushstrokes since a single brushstroke can change colour in an unnatural way. In comparison, when using full style transfer and colour matching, the output image really consists of strokes which are blotches of paint, not just variations of light and dark. For a more detailed discussion of colour preservation in Neural Style Transfer we refer the reader to the Supplementary Material, section 2.1.

\section{Scale Control}\label{sec:ScaleControl}

\begin{figure}[h!]
\includegraphics[width=1\linewidth]{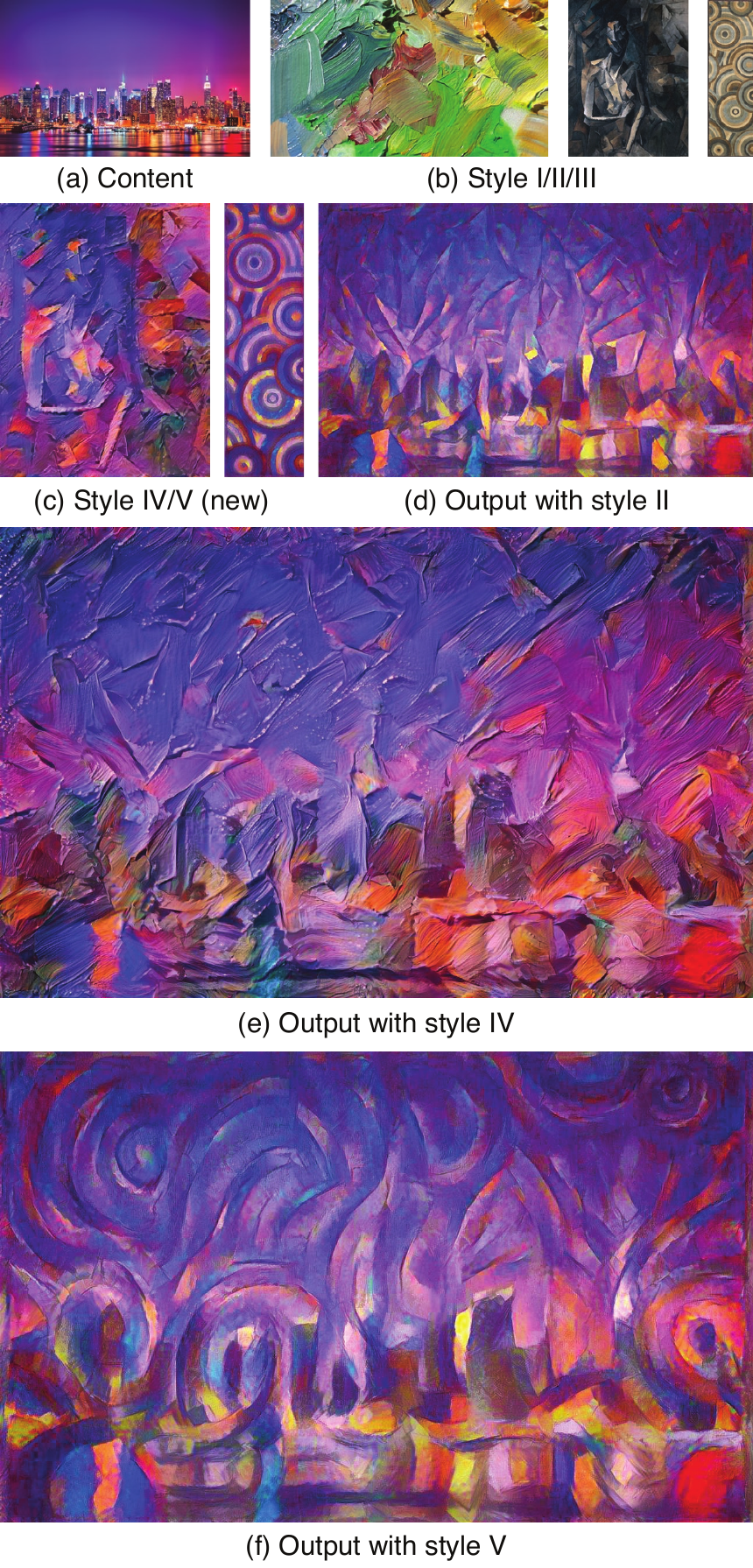}
\caption{Scale control in Neural Style Transfer.
\textbf{(a)} Content image.
\textbf{(b)} Collection of styles used. Style I has dominant brush strokes on the fine scale. Style II has dominant angular shapes on the coarse scale. Style III has dominant round shapes on the coarse scale.
\textbf{(c)} New styles obtained from combining coarse and fine scales of existing styles. Style IV combines fine scale of Style I with coarse scale of Style II. Style V combines fine scale of Style II with coarse scale of Style III.
\textbf{(d)} Output using original Style II.
\textbf{(e)} Output using the new Style IV.
\textbf{(f)} Output using the new Style V. 
All stylisations preserve the colour of the photograph using the colour matching method described in section \ref{sec:ColourMatching}
}
\label{fig:ScaleControl}
\end{figure}

In this section, we describe methods for mixing different styles at different scales and efficiently generating high-resolution output with style at desired scales.

\subsection{Scale control for style mixing}\label{sec:ScaleControlMix}
First we introduce a method to control the stylisation independently on different spatial scales. Our goal is to pick separate styles for different scales. For example, we want to combine the fine-scale brushstrokes of one painting (Fig.~\ref{fig:ScaleControl}(b), Style I) with the coarse-scale angular geometric shapes of another image (Fig.~\ref{fig:ScaleControl}(b), Style II).

We define the style of an image at a certain scale as the distribution of image structures in image neighbourhoods of a certain size $f$.  
In that sense, the colour separation introduced in the previous section can be thought of a special case of scale separation, since image colours are ``structures'' on one-pixel neighbourhoods.
To model image style on larger scales, we use the Gram Matrices from different layers in the CNN.
In particular, a Gram Matrix at layer $\ell$ represents the second-order statistics of image neighbourhoods of size corresponding to the receptive field size $f_\ell$.

Unfortunately, this representation is not factorised over scale. In general, a Gram Matrix $\bG_{\ell}(\bx)$ at a given spatial scale also captures much of the image information on smaller spatial scales and thus shares a lot of information with the Gram Matrix $\bG_{\ell-k}(\bx)$ at a lower layer in the CNN (see Supplementary Material, section 3.1 for more details). 
Therefore, simply combining Gram Matrices from different scales of different images does not give independent control over the different scales.

Here we show a way to combine scales that avoids this problem. We first create a new style image that combines fine-scale information from one image with coarse scale information from another (Fig.~\ref{fig:ScaleControl}(c)). We then use the new style image in the original Neural Style Transfer.
We do this by applying Neural Style Transfer from the fine-scale style image to the coarse-scale style image, using only the Gram Matrices from lower layers in the CNN (e.g., only layer ``conv1\_1'' and ``conv2\_1'' in Fig.~\ref{fig:ScaleControl}). We initialise the optimisation procedure with the coarse-style image and omit the content loss entirely, so that the fine-scale texture from the coarse-style image will be fully replaced. This is based on the observation that the optimisation leaves images structures intact when they are of larger scale than the style features. While this is not guaranteed, as it depends on the optimiser, we empirically find it to be effective for the L-BFGS method typically used in Neural Style Transfer.
The resulting images (Fig.~\ref{fig:ScaleControl}(c)) are used as the input to the original Neural Style Transfer to generate a new stylisations of the cityscape photograph. For example, we combine the fine scale of Style I with the coarse scale of Style II to re-paint the angular cubistic shapes in Fig.~\ref{fig:ScaleControl}(d) with pronounced brushstrokes (Fig.~\ref{fig:ScaleControl}(e)).
Or we combine the fine scale of Style II with the coarse scale of Style III to replace the angular shapes by round structures, giving the image a completely different ``feel'' (compare Fig.~\ref{fig:ScaleControl}(d) with Fig.~\ref{fig:ScaleControl}(f)).

This method enables the creation of a large set of perceptually appealing, new styles by recombining existing ones in a principled way. It also allows for interesting new ways to interpolate between styles by interpolating across spatial scales. For more examples of new styles and results of interpolating between styles, we refer the reader to the Supplementary Material, sections 3.2 and 3.3.

\subsection{Scale control for efficient high resolution}
\begin{figure}
\includegraphics[width=1\linewidth]{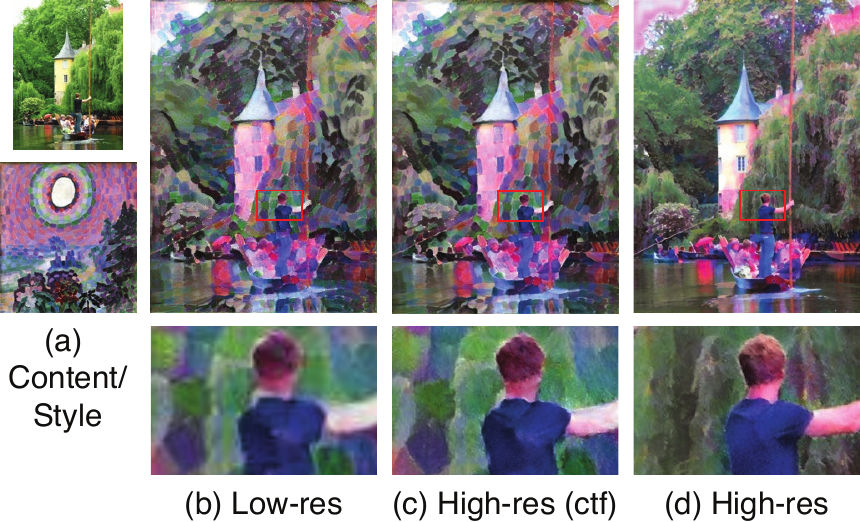}
\caption{Neural Style Transfer in high resolution. 
\textbf{(a)} Content and style images.
\textbf{(b)} Output in low-resolution with total number of pixels equal to $450^2$
\textbf{(c)} Output in high-resolution generated in a coarse-to-fine fashion from \textbf{(b)}.
\textbf{(d)} Output in high-resolution without coarse-to-fine procedure. For both high-resolution images the total number of pixels is $3000^2$ and they can be found in the Supplemental Material.
}
\label{fig:HighRes}
\end{figure}

The existing Neural Style Transfer method does not work well for high-resolution outputs.
Since the receptive fields in a CNN have a fixed size, the stylisation outcome depends on the resolution of the input images: stylisation happens only up to the scale of the receptive fields in the output. 
In practice, we find that for the VGG-19 network, there is a sweet spot around $500^2$ pixels for the size of the input images, such that the stylisation is appealing but the content is well-preserved (Fig.~\ref{fig:HighRes}(b)). For a high-resolution image, however, the receptive fields are typically very small compared to the image, and so only very small-scale structures are stylised (Fig.~\ref{fig:HighRes} (d)).  

Here we show that the same scale separation principle from the previous section can be used in order to produce high-resolution outputs with large-scale stylisation. 
We are given high-resolution content and style images $\bx_C$ and $\bx_S$, both having the same size with  $N^2$ pixels in total. 
We down-sample each image by a factor $k$ such that $N/k$ corresponds to the desired stylisation resolution, e.g., $500^2$ for VGG, and then perform stylisation. The output is now low-resolution of size $N/k$. We can then produce high-resolution output from this image by up-sampling the low-resolution output to $N^2$ pixels, and use this as initialisation for Neural Style Transfer with the original input images $\bx_C$ and $\bx_S$. The style features now capture and can fill-in the high-resolution information from the style image while leaving the coarse-scale stylisation intact (Fig.~\ref{fig:HighRes}(c)).

\begin{figure}
\includegraphics[width=1\linewidth]{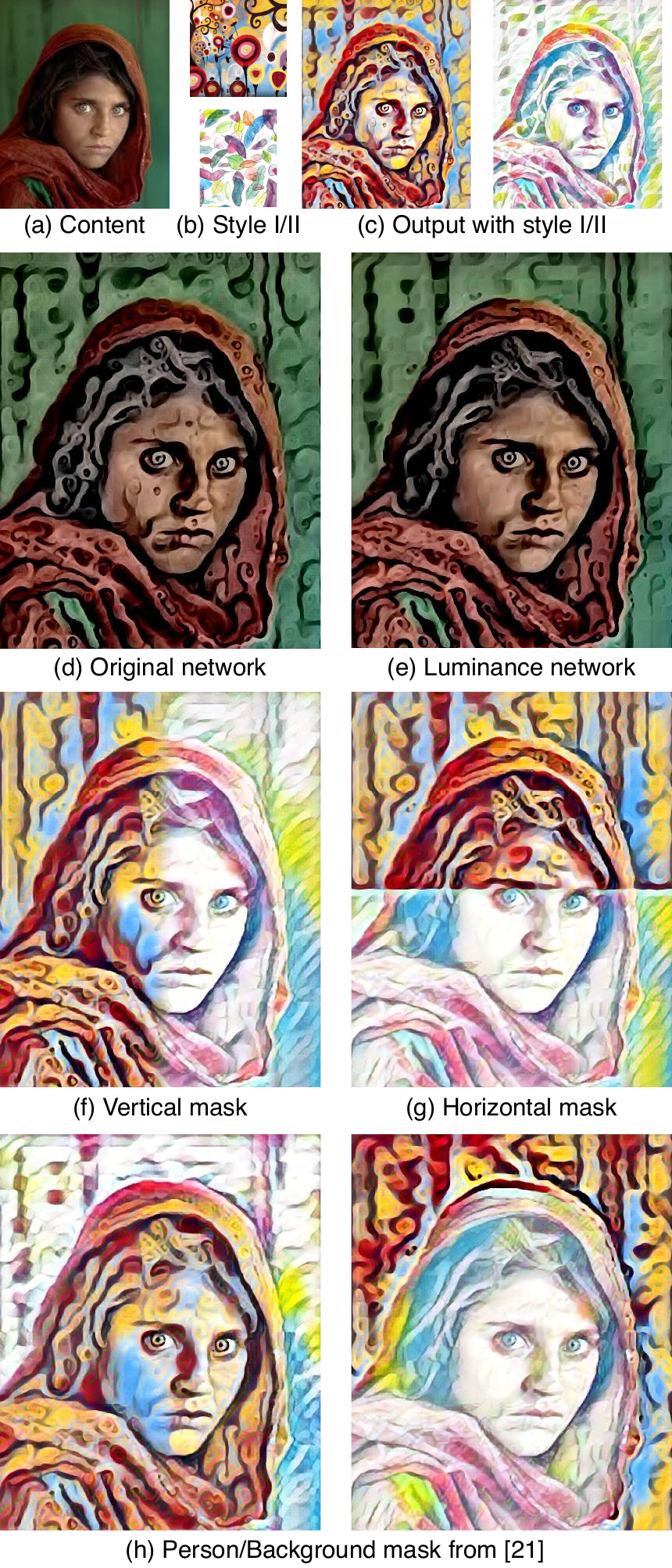}
\caption{Colour and spatial control in Fast Neural Style Transfer.
\textbf{(a)} Content image.
\textbf{(b)} Styles \textit{Candy} and \textit{Feathers}.
\textbf{(c)} Outputs from \cite{johnson_perceptual_2016}, trained with styles shown in \textbf{(b)}.
\textbf{(d)} Simple colour preservation. Luminance channel from \textbf{(c)} is combined with colour channels from \textbf{(a)}.
\textbf{(e)} Colour preservation with luminance network. Output from luminance network is combined with colour channels from \textbf{(a)}.
\textbf{(f)} Vertical separation of styles.  
\textbf{(g)} Horizontal separation of styles.
\textbf{(h)} Separation of styles into person and background using \cite{shen_automatic_2016}.
}
\label{fig:FastNeuralStyle}
\end{figure}

This coarse-to-fine procedure has the additional advantage of requiring fewer iterations in the high-resolution optimisation and thus increasing efficiency. In our experiments we used $2.5$ times fewer iterations for the high-resolution optimisation.
We also noticed that this technique effectively removes low-level noise that is typical for neural image synthesis. In fact, all figures shown in this paper, except for Fig.~\ref{fig:FastNeuralStyle}, were enhanced to high-resolution in that way. The low/high-resolution pairs can be found in the Supplement. Applying this technique iteratively also enables the generation of very high-resolution images that is only limited by the size of the input images and available memory.

\section{Controlling Fast Neural Style Transfer}

A major drawback of Neural Style Transfer is that image generation is relatively slow. Recently, a number of works have shown that one can train a feed-forward CNN to perform stylisation \cite{johnson_perceptual_2016, ulyanov_texture_2016, liWang2016Precomputed}. 
We now show how to apply the spatial and colour control described above to these Fast Neural Style Transfer methods.  
Applying scale control to Fast Neural Style Transfer is trivial, as it entails simply training on the new style image that combines multiple scales.
We use Johnson's excellent publicly-available implementation of Fast Neural Style Transfer \cite{johnson_perceptual_2016}\footnote{github.com/jcjohnson/fast-neural-style}. The networks we train all use the well-tuned default parameters in that implementation including Instance Normalization \cite{ulyanov_instance_2016} (for details see Supplementary Material, section 4). For comparability and to stay in the domain of styles that give good results with Fast Neural Style Transfer, we use the styles published with that implementation.  

\subsection{Colour control}
The simplest way to preserve the colour of the input image is to just use an existing feed-forward stylisation network \cite{johnson_perceptual_2016}, and then combine the luminance channel of the stylisation with the colour channels of the content image (Fig.~\ref{fig:FastNeuralStyle}(c)).
An alternative is to train the feed-forward network exclusively with the luminance channel of the style and content images.
This network then produces a luminance image that can be combined with the colour channels from the input content image (Fig.~\ref{fig:FastNeuralStyle}(d)). For both methods we match the mean luminance of the output image to that of the content image. In general, we find that colour preservation with the luminance network better combines stylisation with structures in the content image
(Fig.~\ref{fig:FastNeuralStyle}(c),(d)).

\subsection{Spatial control}
We now describe training a feed-forward network to apply different styles to different regions.
We show that this can be done with a surprisingly small modification to Johnson's training procedure \cite{johnson_perceptual_2016}, which we illustrate with the following example. 
We create the style image by vertically concatenating the \textit{Candy} and \textit{Feathers} images shown in Fig.~\ref{fig:FastNeuralStyle}(b). Two additional binary guidance channels are added to the style image, i.e., one for the top of the image and one for the bottom.
The style loss function is based on the guided Gram Matrices (Eq.~\ref{eq:guided_loss}).
During training, the feed-forward network takes as input the content image and two guidance channels. The input guidance channels are passed to the loss network to evaluate the spatially-guided losses. Surprisingly, we find that the guidance channels can be kept constant during training: during training we required the feed-forward network to always stylise the lower half of the image with one style and the upper half with another. However, the network robustly learns the correspondence between guidance channels and styles, so that at test time we can pass arbitrary masks to the feed-forward network to spatially guide the stylisation (Fig.~\ref{fig:FastNeuralStyle}(f)-(h)).
By providing an automatically-generated figure-ground segmentation \cite{shen_automatic_2016} we can create an algorithm that performs fast spatially-varying stylisation automatically.
(Fig.~\ref{fig:FastNeuralStyle}(g),(h))

\section{Discussion}
In this work, we introduce intuitive ways to control Neural Style Transfer. We hypothesise that image style includes factors of space, colour, and scale, and present ways to access these factors during stylisation to substantially improve the quality and flexibility of the existing method.

One application of the control methods we present is to combine styles in an interpretable fashion. 
This contrasts with the alternative approach of combining styles by linearly interpolating in the style representation as, for example, is done in the concurrent work of Dumoulin et al.~\cite{dumoulin}. A possible concern with that approach is that if the directions in the style representation do not correspond to perceptual variables, it becomes difficult to generate appealing new styles.
Still, even with our methods the selection of which inputs to combine for aesthetically pleasing results can be challenging. An exciting open research question is to predict what combinations of styles will combine nicely into new, perceptually pleasing styles.

Neural Style Transfer is particularly appealing because it can create new image structures based on the source images. This flexibility arises from the representation of style in terms of spatial summary statistics, in contrast to patch-based methods \cite{image-analogies, ramanarayan-bala, stylit}. However, because it is not clear how the perceptual aspects of style are represented in the summary statistics, it is hard to achieve meaningful \emph{parametric} control over the stylisation. For that it may be necessary to encourage appropriate factorisations of the CNN representations during network training, for example, to learn representations that factorise the image information over spatial scales.
In fact, this touches a fundamental research question in machine vision: to obtain interpretable yet powerful image representations that decompose images into the independent factors of human visual perception.

{\small
\bibliographystyle{ieee}
\bibliography{stylisation}
}

\end{document}